\documentclass[journal]{IEEEtran}

\usepackage{cite}
\usepackage{booktabs}
\usepackage{amsmath,amssymb,amsfonts}
\usepackage{algorithmic}
\usepackage{graphicx}
\usepackage{textcomp}
\usepackage{comment}
\usepackage{makecell}
\usepackage{multirow}
\usepackage{amsmath}
\usepackage[table,xcdraw]{xcolor}
\usepackage[colorlinks=true,allcolors=blue]{hyperref}
\def\BibTeX{{\rm B\kern-.05em{\sc i\kern-.025em b}\kern-.08em
    T\kern-.1667em\lower.7ex\hbox{E}\kern-.125emX}}
\ifCLASSINFOpdf
\else
\fi

\begin{document}
\title{\textit{JammingSnake}: A follow-the-leader continuum robot with variable stiffness based on fiber jamming}

\author{Chen~Qian, \IEEEmembership{Student Member, IEEE}, Tangyou~Liu, \IEEEmembership{Student Member, IEEE}, and~Liao~Wu$^*$, \IEEEmembership{Member, IEEE}
%\and
%\IEEEauthorblockN{2\textsuperscript{nd} Given Name Surname}
%\IEEEauthorblockA{\textit{dept. name of organization (of Aff.)} \\
%\textit{name of organization (of Aff.)}\\
%City, Country \\
%email address or ORCID}
\thanks{This work was supported by the Australian Research Council under Grant DP250102489.}
\thanks{{$^*$}Corresponding author: Liao Wu, {\tt\small dr.liao.wu@ieee.org}.}
\thanks{C. Qian, T. Liu, and L. Wu are with the School of Mechanical and Manufacturing Engineering, University of New South Wales, Sydney, NSW 2052, Australia. }
}

% make the title area
\maketitle

% As a general rule, do not put math, special symbols or citations
% in the abstract or keywords.
\begin{abstract}
Follow-the-leader (FTL) motion is essential for continuum robots operating in fragile and confined environments. 
It allows the robot to exert minimal force on its surroundings, reducing the risk of damage. 
This paper presents a novel design of a snake-like robot capable of achieving FTL motion by integrating fiber jamming modules (FJMs).
The proposed robot can dynamically adjust its stiffness during propagation and interaction with the environment. 
An algorithm is developed to independently control the tendon and FJM insertion movements, allowing the robot to maintain its shape while minimizing the forces exerted on surrounding structures.
To validate the proposed design, comparative tests were conducted between a traditional tendon-driven robot and the novel design under different configurations. 
The results demonstrate that our design relies significantly less on contact with the surroundings to maintain its shape.
This highlights its potential for safer and more effective operations in delicate environments, such as minimally invasive surgery (MIS) or industrial in-situ inspection.
A video demonstration of performance test on \textit{JammingSnake} can be found at \href{https://youtu.be/BHsRPmdTNnw}{https://youtu.be/P2YTCPWHNfo}.

%We believe this robot is capable of XXX with high potential in the future to enhance its payload and accuracy.

\end{abstract}

% Note that keywords are not normally used for peer-reviewed papers.

\IEEEpeerreviewmaketitle

\begin{figure*}[!h]
    \centering
    \includegraphics[width=0.92\textwidth]{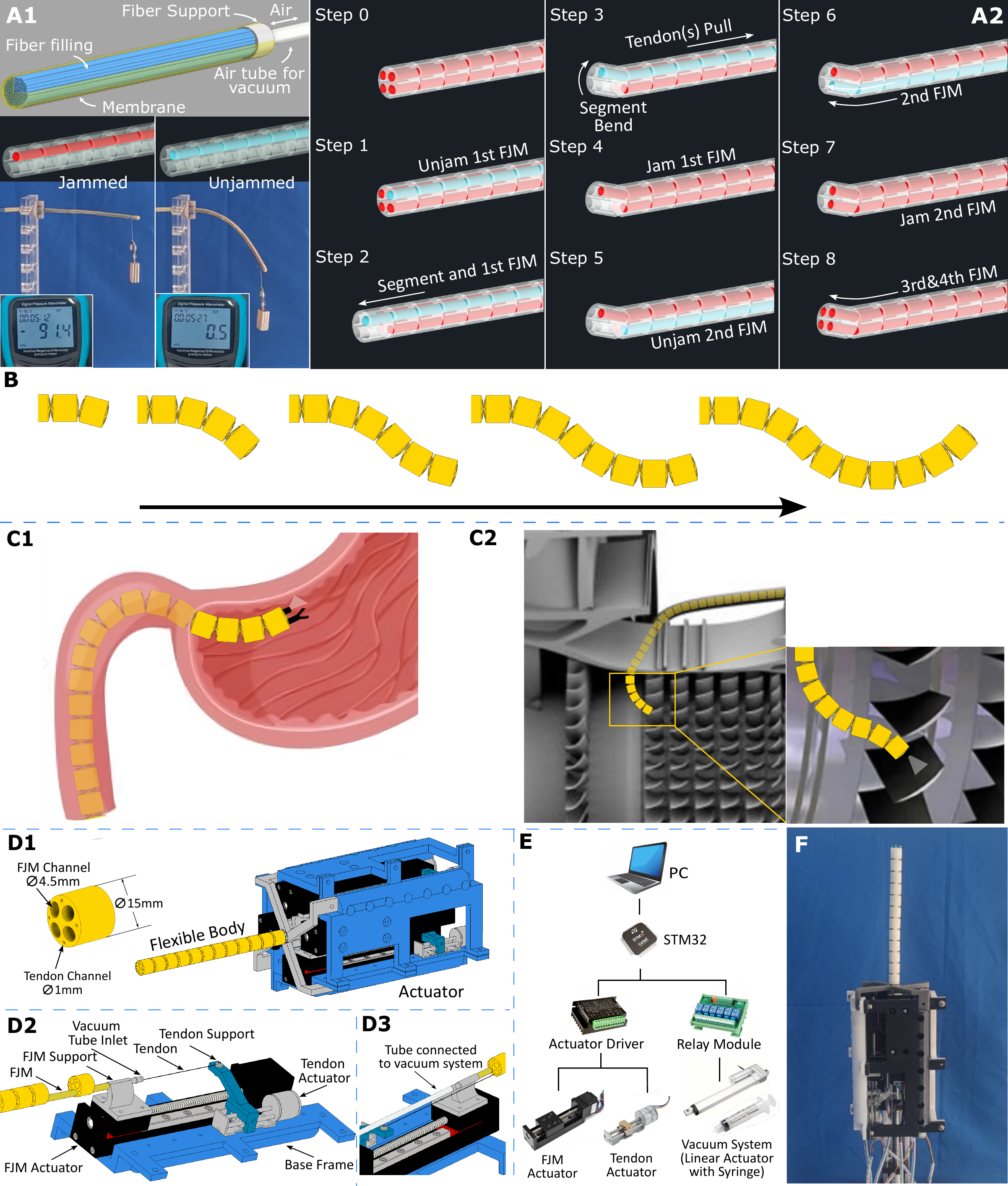}
    \caption{
    \textbf{(A1)}: A Fiber Jamming Module (FJM) is a soft robotic component consisting of fibers within an elastic membrane, capable of achieving rapid and significant stiffness variation through air vacuuming. 
    Photos illustrate the stiffness of an FJM under two conditions: when vacuumed at -90 kPa and when not vacuumed. 
    \textbf{(A2)}: An illustrative demonstration of a complete cycle of our prototype shows the transition from one preserved shape to the next. The procedural details of the FTL mechanism are as follows:
    \textbf{Step 0}: Initially, all FJMs are jammed to preserve the flexible body's shape.
    \textbf{Step 1}: Unjam the first FJM. 
    \textbf{Step 2}: Propagate the robot segments with the unjammed FJM. 
    \textbf{Step 3}: Engage the tendon(s) to steer the head segment to the target orientation.
    \textbf{Step 4}: Jam the propagated FJM to conserve the flexible body's new shape    \textbf{Step 5}: Unjam the second FJM.
    \textbf{Step 6}: Extend the unjammed FJM to the same length as the first.
    \textbf{Step 7}: Jam this FJM to stabilize the robot in its updated configuration.
    \textbf{Step 8}: Repeat Steps 5-7 for the third and fourth FJMs. The flexible body is now in a new stable position (as in Step 0) and ready for the next cycle.
    \textbf{(B)} An illustration demonstrates how FTL motion is facilitated by repeating the cycle in (A).
    \textbf{(C)} Two potential applications for our prototype, which require the flexible body to travel through narrow channels to reach open spaces: medical surgery and engine in-situ inspection.
    \textbf{(D)} Illustration of prototype design, including the flexible body and actuator system. 
    \textbf{(D2)} The actuator section is composed of four identical parts, each equipped with individual actuators to control the pulling force of the tendons and the movement of the FJMs.
    \textbf{(D3)} The back view displays the configuration of the tendons and vacuum tube.
    \textbf{(E)} Illustration of the actuation interface setup.
    \textbf{(F)} The photo of the robot prototype.
    }
    \label{Robot Design}
\end{figure*}

\section{Introduction}

Follow-the-leader (FTL) motion is essential for continuum robots propagating in confined or sensitive environments, as it ensures that the robot body follows the same path traced by its leading section. 
This minimizes the risk of collision and damage to surrounding structures, which is especially important in delicate operations like minimally invasive surgery or industrial inspection of narrow passages \cite{choset1999follow,culmone2021follow,li2017kinematic}. 
The FTL approach allows for precise movement, maintaining low deviation from the desired trajectory, and quick adaptation to environmental changes.

Different approaches have been proposed to address the FTL problem, including algorithmic approximation. Tendon-driven continuum robots (TDCRs) with multiple sections can approximate FTL by controlling the bending angle of each section using carefully designed strategies \cite{gao2019continuum}. However, such designs introduce errors during FTL motion propagation, particularly for trajectories with non-constant curvature segments. Additionally, the fixed length of each section and its ability to bend in only one degree of freedom limit the design’s ability to continuously adapt its shape during propagation for precise FTL motion \cite{grassmann2022fas}.

Acknowledging the shortcomings of algorithmic approximation, another group of approaches aims to achieve ``true" FTL motion through innovative hardware designs. For example, concentric tube continuum robots (CTCRs), which are composed of pre-curved, flexible tubes nested within one another, are capable of FTL motion \cite{gilbert2015concentric}. By rotating and translating the tubes separately, the robot's tip follows a defined trajectory, with the body mirroring the same path. This continuous curvature control ensures precise motion along narrow or complex pathways. However, the pre-bent shapes of the tubes limit the range of operation as the robot's motion is constrained by the fixed curvature of the tubes, reducing adaptability to trajectories outside the predefined curvature. \cite{bergeles2015concentric}. 

To allow a wider operational range, one popular method is to introduce variable stiffness modules \cite{zhong2020novel}. Such a design usually consists of multiple components that can transition between soft and rigid phases \cite{manti2016stiffening,shang2022review}. By steering and propagating the robot in the soft phase and conserving the path's shape in the rigid phase, the robot can potentially enable FTL \cite{degani2006highly}.

One of the earliest attempts, the Highly Articulated Robotic Probe (HARP) by Degani et al. \cite{degani2006highly}, features concentrically assembled continuum robots with stiffness control via segment locking using tendon-pulling forces. However, using the same mechanism for both stiffness variation and steering leads to unintended bending, posture deviations, and increased operational errors in complex environments. Maintaining stiffness in HARP becomes more challenging as the robot lengthens, especially when used as an end effector for a long passive endoscope. Additionally, the large contact area between segments required for sufficient friction limits the robot’s bending flexibility.

More recent designs aimed to focus the tendon locking specifically on robot control while incorporating new mechanisms to enable variations in stiffness, such as probe clamp triggered by Shape Memory Alloy (SMA) \cite{kang2016first,yang2020geometric,wang2022design}. Such designs eliminate the coupling issue and hence reduce operation errors. However, the complexity of the clamp prevents the robot from being downsized to less than 30~mm in diameter. Moreover, the response time of SMA is typically 10 seconds or more. This delayed responsiveness restricts its effectiveness in dynamic environments where rapid actuation or quick adaptation is required, posing a significant barrier to its broader use in real-time robotic systems. 

Similarly, attempts involving Low Melting Point Alloy (LMPA) \cite{mao2024magnetic} have also faced difficulties in practical applications due to the slow response time of the material. The heating time for LMPA is 5-10 seconds, and the cooling time is around 15 seconds.

Jamming modules offer advantages in response time \cite{wang2018development}, shape adaptability \cite{li2019flexible}, and structural simplicity \cite{yang2020geometric,wang2022design}, highlighting their untapped potential in FTL applications \cite{culmone2021follow}. A typical jamming module consists of a sealed soft membrane with internal fillings (Fig.~\ref{Robot Design}A1). When unpressurized, it remains soft, but under high vacuum pressure, the fillings jam together, increasing friction and making the module rigid \cite{fitzgerald2020review}. Based on filling geometry, jamming modules are categorized as granular, fiber, or layer jamming. While granular jamming has been explored in snake-like robots \cite{jiang2014roboticgranular}, fiber jamming remains unexplored. Its ability to restrict longitudinal shape changes while allowing flexibility in other directions makes it particularly suited for continuum robots.

In our previous paper \cite{qian2024effects}, we theoretically and experimentally studied fiber jamming modules (FJMs) for stiffness variation. To assess their integration into slender robots, we designed and fabricated FJMs with a 4mm inner diameter. The experiments identified optimal configurations for achieving the desired characteristics, confirming the feasibility of small-sized FJMs. Meanwhile, factors like fiber bundle arrangement and geometry studied in previous research may be less relevant. However, while FTL was the motivation for the study, no detailed design or implementation of FTL has been investigated. 

Extending \cite{qian2024effects}, the contribution of this paper lies in the design, development, and evaluation of \textit{JammingSnake} - a robot capable of FTL motion based on the strong potential of FJMs.
%To demonstrate the strong potential of FJMs to achieve FTL motion when integrated into a continuum robot, we implemented the design concept from our previous paper and developed the \textit{JammingSnake}—a novel design enabling FTL motion without complex control algorithms. 
The prototype also serves as a promising end effector for endoscopes, addressing stiffness variation challenges as endoscope length increases. Furthermore, it demonstrates strong applicability in confined-space operations, such as medical endoscopy and engine in-situ inspection.

\section{Prototype Design}

\textit{JammingSnake} contains three main components: the robot, the actuator, and the actuation interface  (Fig.~\ref{Robot Design}D). The robot consists of 12 segments, each measuring 15~mm in diameter and 15~mm in length, for a total length of 180~mm. Each segment has four tendon channels and four FJM channels. The tendon channels are 1~mm in diameter. The diameter of the FJM channels is set to 4.5 mm to minimize the gap between the channel and the jammed FJM while preventing excessive friction.

\subsection{Design Concept}

To achieve FTL motion without relying on environmental constraints, a continuum robot must perform three key sub-functions: \textbf{conservation, steering, and propagation} \cite{culmone2021follow}. These functions are executed by alternately jamming and unjamming multiple FJMs integrated into the prototype. By modulating the stiffness of each FJM, the robot dynamically adjusts its structure to maintain shape, steer along a designated path, and move forward. This approach enables precise, adaptable motion, enhancing functionality in diverse and unstructured environments.

Fig.\ref{Robot Design}A2 illustrates a full cycle of the FTL mechanism in \textit{JammingSnake}, where unjammed (soft) FJMs enable steering and propagation while jammed (rigid) FJMs maintain trajectory. Repeating this cycle allows the robot to achieve FTL without environmental support (Fig.\ref{Robot Design}B). While two FJMs can achieve FTL, our prototype integrates four to enhance stability and functionality. In practice, the robot can extend multiple segments per cycle, adapting its configuration to navigate complex paths more effectively.

Additionally, since a minimum of two FJMs is sufficient to achieve FTL motion, the number of FJMs in the prototype can be reduced based on the operational payload. The spare FJMs can either be employed to enhance the stiffness of specific sections of the robot or replaced with alternative components for specialized tasks. This adaptability optimizes the robot’s performance while allowing for efficient resource utilization.

\subsection{Application}

We designed this robot with specific scenarios in mind, aiming to make it highly effective for applications that involve navigating through constrained pathways and achieving precise maneuverability. In these scenarios, the robot must first traverse a long, narrow channel and then enter a larger open space. Once in the open area, the robot needs to independently reach its target without relying on nearby surfaces for support. 

To achieve this, our robot is designed to extend the length of its flexible body, ensuring precision and flexibility while minimizing reliance on complex support mechanisms. FTL motion is particularly crucial in scenarios requiring high agility and stability within large or expansive areas.

This working scenario has applications in medical procedures \cite{liu2024review,razjigaev2023optimal,razjigaev2022end} and industrial inspection \cite{troncoso2022continuum}. For instance, in gastric surgery, an endoscope navigates through the rectum to reach the stomach for precise operations (Fig.~\ref{Robot Design}C1). Similarly, a borescope used for in-situ inspection, such as checking components within a compressor (Fig.~\ref{Robot Design}C2), benefits greatly from the FTL motion to avoid damage due to unnecessary contact. 

\subsection{Actuator Design}

The actuator system measures approximately 220~mm × 120~mm × 120~mm and consists of four identical parts arranged in a circular layout at 90° intervals (Fig.~\ref{Robot Design}D2). Each part includes a base frame that supports a 20~mm stepper motor for tendon pulling and a 28~mm stepper motor for FJM movement. The tendon stepper motors are controlled to bend the head segment (Step~3 in Fig.~\ref{Robot Design}A2), while the FJM stepper motors independently control the insertion or retreat of the FJMs into the flexible body, enabling the desired FTL motion (Step~6). A 50~ml syringe, connected to the linear actuator, serves as a pump to generate a vacuum within the FJM. Each linear actuator is equipped with a relay module that enables bidirectional control, allowing the syringe to alternate between vacuuming and releasing the FJMs as needed. 

A 40~mm base stepper motor drives the robot's insertion, synchronized with the FJM stepper motors. As the base motor advances the robot while the FJM motors retract at the same speed, the flexible body moves forward without changing the FJM insertion depth (Step~2). This motor acts as the robot's proximal section and can be replaced by other actuators, such as a robotic arm, in real applications.

\subsection{Actuation Interface}
The actuation interface (Fig.~\ref{Robot Design}E) uses a 24V supply to power stepper motor drivers and linear actuators, with an STM32 microcontroller handling signal routing and PC communication. This hardware layer provides basic motor control and position monitoring without implementing closed-loop control algorithms. The force modulation demonstrated in our experiments results primarily from the mechanical design's passive compliance rather than active electronic control.

\section{Kinetic Model}\label{Kinetic Model}

In our prototype, tendons are used solely to bend the tip, allowing this actuation method to be easily replaced for future applications. During other phases, the tendons remain loose to avoid obstructing shape preservation. Precise length adjustments for both tendons and FJMs are essential, based on the steering angle and joint dimensions, to ensure optimal shape conservation \cite{kim2013stiffness,hu2022static,huang2021static}.

\begin{figure}[!t]
    \centering
    \includegraphics[width=0.47\textwidth]{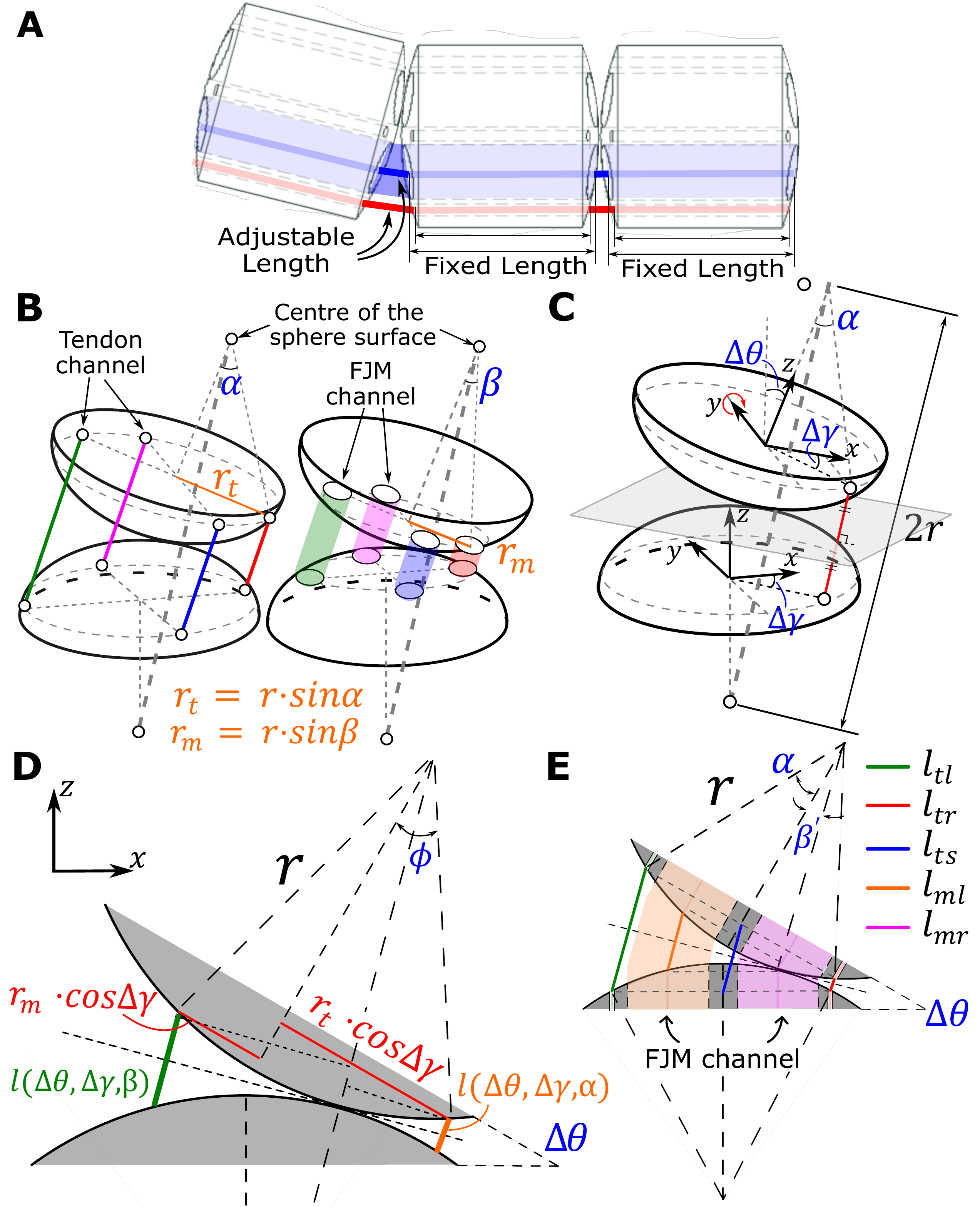}
    \caption{
    An illustration details the calculation of adjusted FJM and tendon lengths:
    \textbf{(A)} Tendon and FJM lengths within each segment are fixed, with adjustable lengths referring between the channel outlets of adjacent segments.
    \textbf{(B)} Robot joint model showing tendon and FJM channel positions, where angles $\alpha$ and $\beta$ represent their positions relative to the sphere's surface symmetry axis.
    \textbf{(C)} Length calculation with a bending degree of $\Delta\theta$ and a twist offset of $\Delta\gamma$ at the tip segment.
    \textbf{(D)} Cross-sectional view in the x-z plane
    \textbf{(E)} Diagram showing the length of tendon and FJM when $\Delta\gamma=0$, where $l_{tl}$: left tendon; $l_{tr}$: right tendon; $l_{ts}$: side tendon; $l_{ml}$: left FJM; $l_{mr}$: right FJM.
    }
    \label{Algorithm}
\end{figure}

Fig.~\ref{Algorithm} illustrates the model used to describe the length variations of tendons and FJMs between two robot segments, accounting for a bending angle of $\Delta\theta$ and the twist offset of $\Delta\gamma$. Considering that the channel outlets for tendons and FJMs are located at different layers within the robot segment (Fig.~\ref{Algorithm}B), the adjustable lengths of the tendons and FJM (Fig.~\ref{Algorithm}C) can be expressed as:
\\

\noindent
\begin{equation}
\begin{aligned}
    &l(\Delta\theta, \Delta\gamma, \psi) 
    = \hspace{0cm} 2r \left( 1 - \frac{\cos\psi}{\cos\phi} \cos\left(\phi - \frac{\Delta\theta}{2}\right) \right) \\
    &where \quad \phi = \tan^{-1}(\tan\psi \cdot \cos\Delta\gamma) \\
    & \quad \quad \quad \;\; \psi =
    \begin{cases} 
    \alpha & \text{for tendon} \\
    \beta & \text{for FJM}
    \end{cases}
\end{aligned}.
\label{l_calc_tendon}
\end{equation}

The FJM diameter is too large to be treated as a wire, so its movement is approximated based on the centerline of the FJM channels. During prototype testing, we simplified the adjusted lengths of tendons and FJMs by setting the right tendon channel to $\Delta\gamma = 0$:

\begin{equation}
\begin{aligned}
    l_{tl}&=l(\Delta\theta,\pi,\alpha)= 2r(1 - \cos(\alpha + \frac{\Delta\theta}{2})) \\
    l_{tr}&=l(\Delta\theta,0,\alpha)= 2r(1 - \cos(\alpha - \frac{\Delta\theta}{2})) \\
    l_{ts}&=l(\Delta\theta,\frac{\pi}{2},\alpha)= 2r(1 - \cos\alpha \cdot \cos\frac{\Delta\theta}{2}) \\
    l_{ml}&=l(\Delta\theta,\frac{\pi}{4},\beta)= 2r(1 - \frac{\cos\beta}{\cos\beta'}\cos(\beta' + \frac{\Delta\theta}{2}) \\
    l_{mr}&=l(\Delta\theta,\frac{3\pi}{4},\beta)= 2r(1 - \frac{\cos\beta}{\cos\beta'}\cos(\beta' - \frac{\Delta\theta}{2})\\
    &where \quad \beta'=tan^{-1}(1/\sqrt{2}cos\beta)
\end{aligned}.
\end{equation}

These calculations help us regulate the tendon pulling force during prototype propagation, preventing the tendon from being overloaded or over-tensioned. Additionally, it enhances the precision of positioning the FJMs relative to the robot segments, which is critical for achieving the best FTL motion. 

\begin{figure*}[t]
    \centering
    \includegraphics[width=0.95\textwidth]{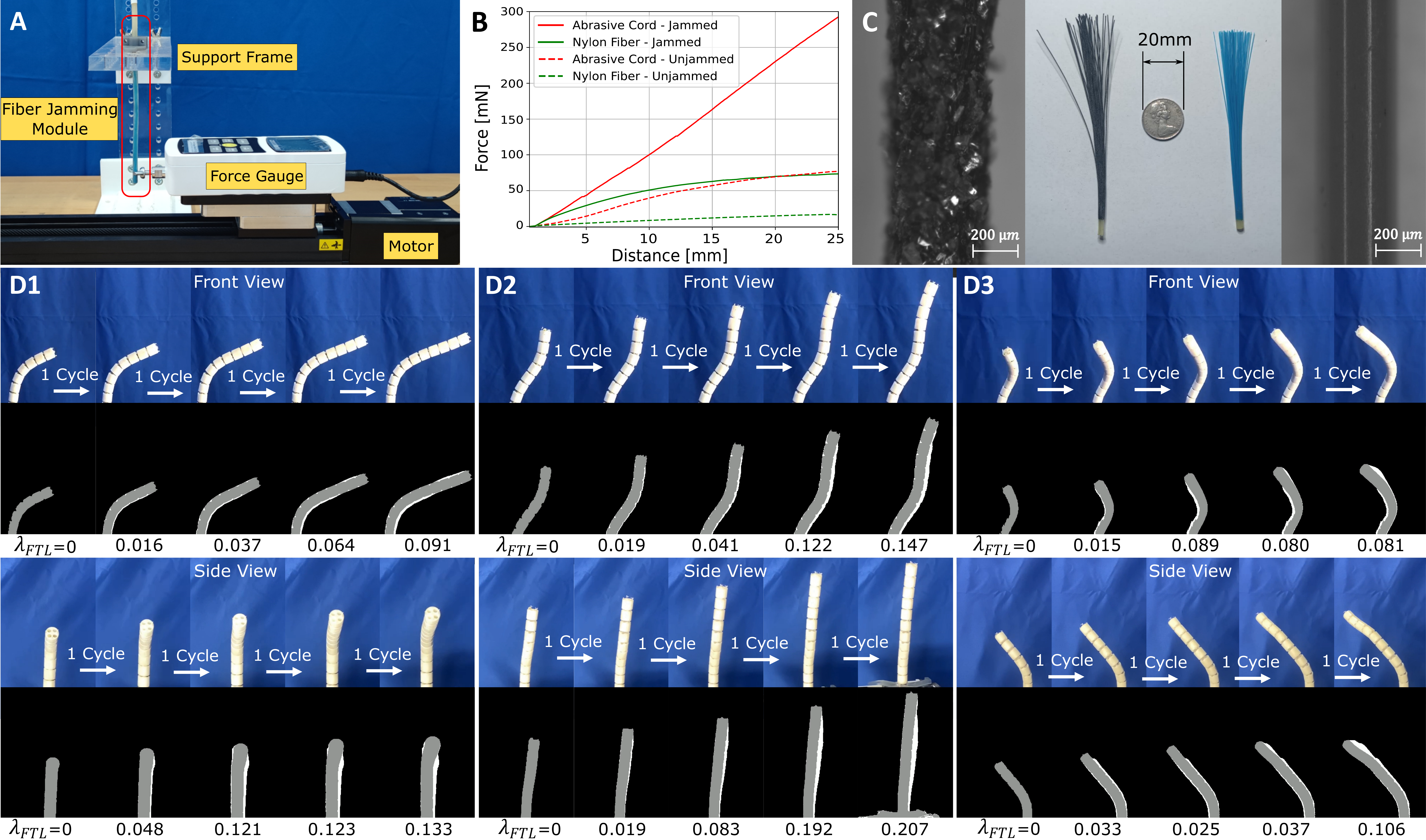}
    \caption{
    \textbf{(A)} The stiffness test setup used a force gauge to measure the FJM’s resistance force and a stepper motor to record displacement.
    \textbf{(B)} Results show that FJMs with abrasive cords have higher, more linear stiffness than nylon fiber bundles in both jammed and unjammed states.
    \textbf{(C)} Images at 50x magnification show that abrasive cords have rougher surfaces than nylon fibers, reducing slippage and increasing jammed stiffness.
    \textbf{(D)} An illustration shows how the prototype maintains its shape while propagating in  C shape \textbf{(D1)}, S shape \textbf{(D2)} and spiral shape \textbf{(D3)} configurations. The value in each figure shows the calculated FTL error as the ratio of the current robot’s area (grey area) to the sweeping area (white area).
    } 
    \label{FJM Stiffness Test}
\end{figure*}

\section{Performance Test}

\subsection{Shape Conservation Performance}
A preliminary test was conducted to evaluate the updated FJM with abrasive cords for shape conservation and to assess its performance during propagation. The FJMs were jammed as the robot segments advanced. 

In our prior research \cite{qian2024effects}, we determined the optimal configuration for the number and density of fibers in FJMs with an inner diameter of 4~mm, using 0.4~mm fibers at a packing density of 56\%, resulting in a stiffness variation of up to 3400\% when no deflection occurs during the rigid phase. Further improvements in this work involved replacing nylon fibers with abrasive cords to enhance overall stiffness, building on insights from previous studies \cite{jadhav2022variable}. Tests confirm that FJMs with abrasive cords demonstrate a stiffness variation ratio comparable to that of nylon fibers while offering increased overall rigidity (Fig.~\ref{FJM Stiffness Test}A\&B). Moreover, high-magnification microscopy (50x) revealed the abrasive cords' rough surface texture, which enables a more linear jamming response and enhances stability during robot propagation (Fig.~\ref{FJM Stiffness Test}C).

To assess the accuracy of the shape conservation behavior of our prototype, we employed a sweeping-area method to visualize the posture error, referencing the work of Grassmann et al. \cite{grassmann2022fas}. A custom script processed video footage of the robot’s propagation, categorizing it into three colors: black for the background, grey for the robot, and white for the area swiped by the robot, which provides a clear visualization of the robot’s movement and spatial occupancy. The posture error $\lambda_{FTL}$ then can be calculated: 

\begin{equation}
    \lambda_{FTL}=\frac{the\:sweeping\:area\:(white\:area)}{the\:current\: robot’s\:area\:(grey\:area)}
\end{equation}

As shown in Fig.~\ref{FJM Stiffness Test}D, the error $\lambda_{FTL}$ ranges from 0.09 to 0.2 after 4 segments extension, while the error of other designs such as HARP \cite{degani2006highly} reaches 0.55 with the same extension operation. This suggests that our prototype demonstrates exceptional performance in maintaining shape conservation while propagating. This finding reinforces the effectiveness of the updated FJM with abrasive cords in achieving the desired operational characteristics in robotic applications. This stability is crucial for retaining the robot’s configuration along complex paths, ensuring consistent positioning in demanding environments.

\begin{figure*}[!h]
    \centering
    \includegraphics[width=0.9\textwidth]{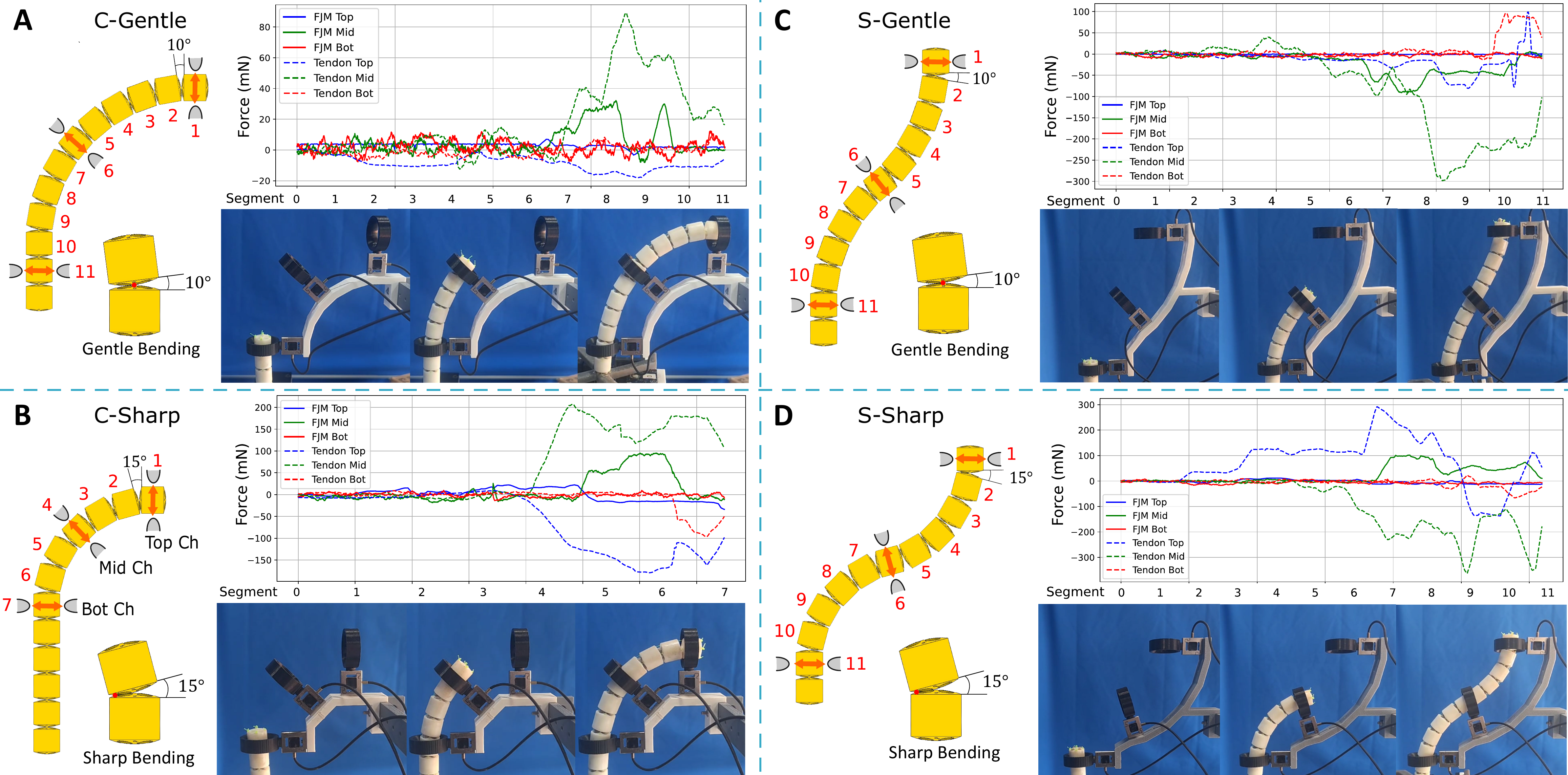}
    \caption{The test included two bending scenarios: gentle bending (10° angle between two adjacent segments) and sharp bending (15° between two adjacent segments, the robot's maximum bending angle). For each scenario, the robot performed one-section (C shape) and two-section (S shape) bending to evaluate shape conservation. Force sensors at three checkpoints (bottom, middle, and top of the trajectory) measured forces exerted by the robot. The graphs compare force responses for TDCR and FJM-assisted operations, showing significantly lower forces exerted by FJM-assisted operations across all scenarios.}
    \label{FTL config}
\end{figure*}

\begin{table*}[t]
\centering
\caption{Summary of force measurements}
\label{force sensor data}
\footnotesize
\setlength{\tabcolsep}{3.5pt}
\begin{tabular}{@{}ll|c|cccc|c|cccc|c|cccc@{}}
\toprule
\multicolumn{2}{c|}{} & \multicolumn{5}{c|}{\textbf{Phase I}} & \multicolumn{5}{c|}{\textbf{Phase II}} & \multicolumn{5}{c}{\textbf{Phase III}} \\
\cmidrule(r){3-7} \cmidrule(r){8-12} \cmidrule(r){13-17}
\multicolumn{2}{c|}{\textbf{Exerted Force (mN)}} & \textbf{Tendon} & \multicolumn{4}{c|}{\textbf{FJM assisted}} & \textbf{Tendon} & \multicolumn{4}{c|}{\textbf{FJM assisted}} & \textbf{Tendon} & \multicolumn{4}{c}{\textbf{FJM assisted}} \\
\cmidrule(lr){3-7} \cmidrule(lr){8-12} \cmidrule(lr){13-17} &
 &  & \textbf{ES1} & \textbf{ES2} & \textbf{ES3} & \textbf{ES4} &  & \textbf{ES1} & \textbf{ES2} & \textbf{ES3} & \textbf{ES4} & & \textbf{ES1} & \textbf{ES2} & \textbf{ES3} & \textbf{ES4} \\
\midrule
C-Gentle & Bottom & \color[HTML]{009901}{NT*} & \color[HTML]{009901}{NT*} & \color[HTML]{009901}{NT*} & \color[HTML]{009901}{NT*} & \color[HTML]{009901}{NT*} & \color[HTML]{009901}{NT*} & \color[HTML]{009901}{NT*} & \color[HTML]{009901}{NT*} & \color[HTML]{009901}{NT*} & \color[HTML]{009901}{NT*} & \color[HTML]{009901}{NT*} & \color[HTML]{009901}{NT*} & \color[HTML]{009901}{NT*} & \color[HTML]{009901}{NT*} & \color[HTML]{009901}{NT*} \\
 & Middle & -- & -- & -- & -- & -- & \color[HTML]{CB0000}{90$\pm$16} & \color[HTML]{009901}{34$\pm$6} & \color[HTML]{009901}{33$\pm$10} & \color[HTML]{009901}{35$\pm$11} & \color[HTML]{009901}{33$\pm$9} & \color[HTML]{CB0000}{67$\pm$17} & \color[HTML]{009901}{NT*} & \color[HTML]{009901}{NT*} & \color[HTML]{009901}{NT*} & \color[HTML]{009901}{NT*} \\
 & Top & -- & -- & -- & -- & -- & -- & -- & -- & -- & -- & \color[HTML]{009901}{NT*} & \color[HTML]{009901}{NT*} & \color[HTML]{009901}{NT*} & \color[HTML]{009901}{NT*} & \color[HTML]{009901}{NT*} \\
\addlinespace
C-Sharp & Bottom & \color[HTML]{009901}{NT*} & \color[HTML]{009901}{NT*} & \color[HTML]{009901}{NT*} & \color[HTML]{009901}{NT*} & \color[HTML]{009901}{NT*} & \color[HTML]{CB0000}{184$\pm$19} & \color[HTML]{009901}{29$\pm$5} & \color[HTML]{009901}{31$\pm$9} & \color[HTML]{009901}{30$\pm$6} & \color[HTML]{009901}{33$\pm$5} & \color[HTML]{CB0000}{176$\pm$19} & \color[HTML]{009901}{NT*} & \color[HTML]{009901}{NT*} & \color[HTML]{009901}{NT*} & \color[HTML]{009901}{NT*} \\
 & Middle & -- & -- & -- & -- & -- & \color[HTML]{CB0000}{172$\pm$10} & \color[HTML]{009901}{78$\pm$8} & \color[HTML]{009901}{73$\pm$11} & \color[HTML]{009901}{78$\pm$14} & \color[HTML]{009901}{81$\pm$13} & \color[HTML]{CB0000}{162$\pm$11} & \color[HTML]{009901}{NT*} & \color[HTML]{009901}{NT*} & \color[HTML]{009901}{NT*} & \color[HTML]{009901}{NT*} \\
 & Top & -- & -- & -- & -- & -- & -- & -- & -- & -- & -- & \color[HTML]{CB0000}{88$\pm$7} & \color[HTML]{009901}{NT*} & \color[HTML]{009901}{NT*} & \color[HTML]{009901}{NT*} & \color[HTML]{009901}{NT*} \\
\addlinespace
S-Gentle & Bottom & \color[HTML]{009901}{NT*} & \color[HTML]{009901}{NT*} & \color[HTML]{009901}{NT*} & \color[HTML]{009901}{NT*} & \color[HTML]{009901}{NT*} & \color[HTML]{CB0000}{101$\pm$21} & \color[HTML]{009901}{NT*} & \color[HTML]{009901}{NT*} & \color[HTML]{009901}{NT*} & \color[HTML]{009901}{NT*} & \color[HTML]{CB0000}{74$\pm$22} & \color[HTML]{009901}{NT*} & \color[HTML]{009901}{NT*} & \color[HTML]{009901}{NT*} & \color[HTML]{009901}{NT*} \\
 & Middle & -- & -- & -- & -- & -- & \color[HTML]{CB0000}{290$\pm$39} & \color[HTML]{009901}{100$\pm$6} & \color[HTML]{009901}{98$\pm$7} & \color[HTML]{009901}{107$\pm$11} & \color[HTML]{009901}{105$\pm$11} & \color[HTML]{CB0000}{227$\pm$9} & \color[HTML]{009901}{21$\pm$4} & \color[HTML]{009901}{17$\pm$2} & \color[HTML]{009901}{17$\pm$3} & \color[HTML]{009901}{18$\pm$1} \\
 & Top & -- & -- & -- & -- & -- & -- & -- & -- & -- & -- & \color[HTML]{CB0000}{84$\pm$12} & \color[HTML]{009901}{NT*} & \color[HTML]{009901}{NT*} & \color[HTML]{009901}{NT*} & \color[HTML]{009901}{NT*} \\
\addlinespace
S-Sharp & Bottom & \color[HTML]{CB0000}{129$\pm$29} & \color[HTML]{009901}{NT*} & \color[HTML]{009901}{NT*} & \color[HTML]{009901}{NT*} & \color[HTML]{009901}{NT*} & \color[HTML]{CB0000}{249$\pm$24} & \color[HTML]{009901}{NT*} & \color[HTML]{009901}{NT*} & \color[HTML]{009901}{NT*} & \color[HTML]{009901}{NT*} & \color[HTML]{CB0000}{144$\pm$25} & \color[HTML]{009901}{NT*} & \color[HTML]{009901}{NT*} & \color[HTML]{009901}{NT*} & \color[HTML]{009901}{NT*} \\
 & Middle & -- & -- & -- & -- & -- & \color[HTML]{CB0000}{418$\pm$40} & \color[HTML]{009901}{103$\pm$5} & \color[HTML]{009901}{107$\pm$18} & \color[HTML]{009901}{108$\pm$10} & \color[HTML]{009901}{111$\pm$12} & \color[HTML]{CB0000}{340$\pm$49} & \color[HTML]{009901}{52$\pm$4} & \color[HTML]{009901}{63$\pm$15} & \color[HTML]{009901}{55$\pm$13} & \color[HTML]{009901}{61$\pm$16} \\
 & Top & -- & -- & -- & -- & -- & -- & -- & -- & -- & -- & \color[HTML]{CB0000}{45$\pm$28} & \color[HTML]{009901}{NT*} & \color[HTML]{009901}{NT*} & \color[HTML]{009901}{NT*} & \color[HTML]{009901}{NT*} \\
\bottomrule
\end{tabular}

\vspace{4pt}
\raggedright
\footnotesize
* The table lists the highest average values per segment after the full pass at three checkpoints. Values are shown as mean $\pm$ standard deviation. \\
* \textbf{ES1 - ES4} represent different extension strategies, each using a different first-forward FJM. \\
* \textbf{NT} = No Touch.

\end{table*}

\begin{table*}[ht]
\centering
\caption{Comparison with follow the leader continuum robot based on stiffness variation mechanisms} 
\label{Robot Comparison Table}
\begin{tabular}{lllllll}

\hline
\textbf{Ref}                                              & \textbf{OD {[}mm{]}} & \textbf{Length {[}mm{]}} & \textbf{Stiffness Variation Type} & \makecell[l]{\textbf{Max Bending}\\ \textbf{Range {[}$^{\circ}${]}}} & \makecell[l]{\textbf{Actuation Time of} \\ \textbf{Stiffness Variation  {[}s{]}}} & \makecell[l]{\textbf{Coupling between steering} \\ \textbf{and stiffness variation}} \\ \hline
This work                                                 & 15                & 180                      & Fiber jamming                     & 180                   & 2s                              & Decoupled                                     \\ \hline
\cite{amanov2021tendon,grassmann2022fas} & 7                    & 165                      & Actuated segment                  & 180                   & N/A*                               & Coupled                                     \\ \hline
\cite{degani2006highly}                  & 12                   & 300                      & Segment locking by tendon         & 15 per segment        & $\approx$1s*                   & Coupled                                     \\ \hline
\cite{bishop2022novel}             & 38                   & 84                       & SMP                               & 70                   & 5                        & Decoupled                                      \\ \hline
\cite{mao2024magnetic}                 & 4x2                   & 400                      & LMPA                              & 180                     & 5 / 15 (heating/cooling)                          & Decoupled                                      \\ \hline
\cite{kang2016first}                     & 30                   & $\approx$240*                     & Rod clamping                      & 180                   & 10-11                               & Decoupled                                      \\ \hline
\multicolumn{7}{l}{\begin{tabular}[c]{@{}l@{}}* Some data in the table for previous designs were not explicitly provided in their respective papers; these values were estimated based on \\ available illustrations and demonstration videos, or labelled as ``N/A" in the table.
\\
\end{tabular}}
\end{tabular}
\end{table*}

\subsection{Force Exertion Performance}

\textit{JammingSnake} highlights minimal force exertion on surrounding structures as a key innovation achieved through FJMs. To evaluate its effectiveness, we developed a testing methodology that simulates narrow paths, allowing us to assess the robot’s performance in constrained environments.

In our robot segment design, the standard bending angle between two segments is set to 10$^{\circ}$, where the segments make contact at the center of their top surfaces. This gentle bend minimizes the offset in the FJM channels, allowing the FJMs to pass through smoothly during propagation.

For sharper bending, the bending angle can reach up to 15$^{\circ}$, bringing the segments into contact at the closer tendon channels. While the FJMs can still pass through at this angle, the increased offset introduces additional friction due to the movement of the FJMs. This added friction can result in greater deformation of the robot from its conserved shape and, in confined areas, can lead to increased force exerted on the surrounding environment. Therefore, sharp bending is tested as an extreme scenario to quantify the maximum force the robot is expected to exert on its surroundings.

The testing procedure involved creating a controlled environment with narrow pathways to simulate real-world constraints. Three checkpoints were placed at the beginning, middle, and end of the pathway, each with a 20~mm inner diameter, while the robot's outer diameter was 15~mm. Single-axis force sensors at each checkpoint recorded the forces exerted by the robot to the surrounding, helping assess its interaction with the environment and providing data on shape maintenance and stability.

The two most common operation scenarios, single-section bending (C-shape) and two-section bending (S-shape), were tested under both gentle and sharp bending conditions. The propagation process was divided into three phases based on the tip position:
\begin{itemize}
    \item Phase I: The tip passes the bottom checkpoint;
    \item Phase II: The tip crosses the middle checkpoint;
    \item Phase III: The tip reaches the top checkpoint and continues until the process is complete.
\end{itemize}

The exerted force at each of the three checkpoints was calculated as the mean force applied by each individual segment over the full duration of the passing process. The highest mean force recorded during each phase at each checkpoint was used as a benchmark for analysis to assess the prototype's shape conservation performance.

To evaluate the potential posture error caused by different extension orders of FJMs due to the asymmetric structure of our prototype, we repeated the tests using four FJMs as the first extension module as an individual extension strategy, followed by the extension of the remaining FJMs in a CW order. For comparative analysis, we conducted the same tests on the prototype without activating the FJMs, allowing it to function as a standard Tendon-Driven Continuum Robot (TDCR).

Before the force exertion tests, we controlled the prototype manually to move forward while monitoring the force sensor readings to minimize exerted force. The prototype was then operated using the optimized routine during the actual tests, with the required forward distance and bending angle determined by the algorithm detailed in Section \ref{Kinetic Model}. Three tests were conducted for each extension strategy, including tendon-only operation. By analyzing the differences in force exertion between the two configurations, we aimed to quantify the benefits of FJMs in terms of shape conservation and overall performance.

Fig.~\ref{FTL config} presents photos of the prototype progressing through the three phases, along with the raw force reading data. The processed data are summarized in Table.~\ref{force sensor data}. Our prototype demonstrates significantly reduced environmental interaction forces compared to conventional TDCR. 

A noteworthy result is that our prototype exerts significantly less force on the surroundings than a standard TDCR of the same diameter. This effect is especially pronounced in C shaped pathways, which are relatively simple to navigate. Whether in a sharp or gentle C shape configuration, our prototype did not apply any force to the bottom checkpoint and exerted less than 94~mN to middle checkpoints the entire propagation.

In contrast, the non-FJM-assisted robot exerted up to 190~mN at the middle checkpoint and 198~mN at the bottom checkpoint. Consequently, in real-world scenarios where pathways are confined, it can be predicted that nearly the entire pathway, except the area near the final target position, will experience significantly higher forces. 

A similar situation was also observed in the S shape configurations with distinct advantages emerging for the FJM approach. In the S-Gentle configuration, the FJM-assisted robot consistently exerts lower forces at all checkpoints, with values at up to 124~mN at the checkpoints compared to the force at a maximum of 320~mN in the Tendon-only setup. The same trend is observed in the S-Sharp configuration, where the FJM-assisted robot demonstrates significantly lower force exertion, recording only 110~mN at the middle checkpoint versus 444~mN in the Tendon-only setup. 

It is noted that the force readings under different extension strategies in the same testing scenarios are similar. This demonstrates that the forwarding order of FJMs has only a minor impact on posture error, resulting in limited effects on the force exerted on the surroundings.

\section{Discussion}

Table.~\ref{Robot Comparison Table} compares our prototype with multiple existing continuum robot designs that also achieve FTL motion through stiffness variation mechanisms. Based on the table comparison, \textit{JammingSnake} shows balanced merits across various aspects. Here are the key points highlighting why our prototype is either better or at least highly competitive as a novel design:

\begin{enumerate}
    \item \textbf{Compact Size}: With an outer diameter (OD) of 15~mm and a length of 180~mm, our design is both compact and appropriately sized for many constrained environments where continuum robots must operate, from medical to industrial applications. Other designs vary significantly in size; for example, the rod clamping mechanism \cite{kang2016first} has a much larger OD of 30~mm, which may limit its applications in narrow or delicate environments. 
    
    \item \textbf{Reasonable Actuation Time of Stiffness Variation}: Our prototype achieves a fast actuation time of 2 seconds, which is significantly competitive. Some other designs, like those using SMP \cite{bishop2022novel} and LMPA \cite{mao2024magnetic}, have actuation times that can be much slower. Our faster response time can be critical for applications that demand real-time adjustments where delays could impact efficiency and safety. In addition, the current actuation time is limited by the actuator, not the mechanism, making it easier to be reduced to less than 0.5~s.
    
    \item \textbf{High Shape Conversation Performance}: 
     Our prototype employs a decoupled control mechanism, avoiding the posture errors seen in other designs during the stiffness variation procedure. Coupled control systems, which link robot movement and stiffness variation through shared actuation like tendon-pulling, often cause unintended interactions such as unintentional vibration of the robot during stiffness adjustments or disrupted stiffness during bending. For instance, HARP \cite{degani2006highly} frequently exhibits end-effector shaking during propagation, worsening as its length increases. Such interactions complicate precise control, increasing posture errors. In contrast, our decoupled mechanism minimizes shape changes during propagation.
\end{enumerate}

Overall, our prototype achieves a competitive balance of compact size, efficient actuation, and exceptional error resilience over distance. Its fiber jamming mechanism delivers a broad, continuous range of stiffness control, enhancing its robustness and suitability for FTL continuum robots. These features make it highly promising for applications requiring precision, adaptability, and reliable performance in confined or hard-to-access environments. 

Although our current prototype has a 15~mm diameter, it could be reduced to 10~mm with smaller FJMs or improved configurations of FJM channels, enhancing compatibility with narrow surgical or industrial inspection channels. In addition, by replacing the current pump system with a bidirectional pump system, we can easily reduce the time of stiffness variation actuation to less than 1 second, which can be compatible with other designs.

\section{Conclusion}
In conclusion, we have developed \textit{JammingSnake}, a continuum robot capable of achieving FTL motion through the use of FJMs, providing significant advantages over existing designs. This approach introduces a unique mechanism where FJMs, capable of transitioning between flexible and rigid states, allow the robot to conserve its shape and adaptively stiffen specific sections as needed. Through this design, we have demonstrated significantly reduced force exertion on surrounding structures compared to traditional TDCRs, highlighting the advantages of FJMs in minimizing friction and improving stability in narrow, confined spaces. Additionally, our design supports precise shape control and minimizes manipulation error across extended distances from the actuator, making it particularly suitable for applications as an end effector in medical and industrial scenarios. Our proposed modifications, including optimized actuation and control systems, lay the groundwork for enhancing the robot’s efficiency, adaptability and maneuverability in complex operational environments.

% use section* for acknowledgment
% \section*{Acknowledgment}

% The authors would like to thank Dr. Shuhua Peng for helping with the stiffness test equipment, A/Prof. Jiangtao Xu for advice on the fabrication of FJM, Dr Yuyan Yu for helping with the magnification photos, and Prof. Wei Gao for advice on the theoretical study of fiber jamming.

% Can use something like this to put references on a page
% by themselves when using endfloat and the captionsoff option.
\ifCLASSOPTIONcaptionsoff
  \newpage
\fi

\bibliographystyle{IEEEtran}
\bibliography{citations}

\end{document}